\documentclass[runningheads]{llncs}
\usepackage{booktabs}
\usepackage{color}%
\usepackage{threeparttable}

\usepackage{amsmath}
\usepackage{color, soul}
\usepackage{lineno,hyperref}
\hypersetup{
	colorlinks,
	linkcolor=[rgb]{1.0, 0.0, 0.0},
	citecolor=[rgb]{0.0, 0.2, 0.8},
	urlcolor =[rgb]{0.0, 0.0, 0.8}
}

\usepackage{adjustbox}

\usepackage{array}
\usepackage{booktabs}
\usepackage[table, dvipsnames, svgnames, x11names]{xcolor}

\usepackage[T1]{fontenc}
\usepackage{graphicx,verbatim}
\begin{document}
%
% \title{VideoPath-LLaVA: Introducing Pathology Video Instruction Tuning for Enhanced Visual Descriptions and Diagnostic Reasoning}
\title{VideoPath-LLaVA: Pathology Diagnostic Reasoning Through Video Instruction Tuning}

\titlerunning{VideoPath-LLaVA}

\begin{comment}  %% Removed for anonymized MICCAI 2025 submission
\author{First Author\inst{1}\orcidID{0000-1111-2222-3333} \and
Second Author\inst{2,3}\orcidID{1111-2222-3333-4444} \and
Third Author\inst{3}\orcidID{2222--3333-4444-5555}}
%
\authorrunning{F. Author et al.}
\institute{Princeton University, Princeton NJ 08544, USA \and
Springer Heidelberg, Tiergartenstr. 17, 69121 Heidelberg, Germany
\email{lncs@springer.com}\\
\url{http://www.springer.com/gp/computer-science/lncs} \and
ABC Institute, Rupert-Karls-University Heidelberg, Heidelberg, Germany\\
\email{\{abc,lncs\}@uni-heidelberg.de}}

\end{comment}

\author{Trinh Vuong and Jin Tae Kwak}  %% Added for anonymized MICCAI 2025 submission
\authorrunning{Trinh et al.}
\institute{Korea University \\
    \email{jkwak@korea.ac.kr}
    }

\maketitle              % typeset the header of the contribution
\begin{abstract}
We present VideoPath-LLaVA, the first large multimodal model (LMM) in computational pathology that integrates three distinct image scenarios, single patch images, automatically keyframe-extracted clips, and manually segmented video pathology images, to mimic the natural diagnostic process of pathologists. By generating detailed histological descriptions and culminating in a definitive sign-out diagnosis, VideoPath-LLaVA bridges visual narratives with diagnostic reasoning.

Central to our approach is the VideoPath-Instruct dataset, comprising 4278 video and diagnosis-specific chain-of-thought instructional pairs sourced from educational histopathology videos on YouTube. 
Although high-quality data is critical for enhancing diagnostic reasoning, its creation is time-intensive and limited in volume. To overcome this challenge, we transfer knowledge from existing single-image instruction datasets to train on weakly annotated, keyframe-extracted clips, followed by fine-tuning on manually segmented videos. VideoPath-LLaVA establishes a new benchmark in pathology video analysis and offers a promising foundation for future AI systems that support clinical decision-making through integrated visual and diagnostic reasoning. Our code, data, and model are publicly available at  \\ \url{https://github.com/trinhvg/VideoPath-LLaVA}.

% This methodology enables the integration of detailed visual descriptions with chain-of-thought diagnostic reasoning, a capability that is critical for mimicking the decision-making process of expert pathologists.

\end{abstract}
%
%
%4036+242
\section{Introduction}
Recent advancements in large language models (LLMs) and large multi-modal models (LMMs) have catalyzed significant improvements in the visual language instruction-tuning process, i.e., supervised fine-tuning (SFT). Reasoning LLMs, in particular, excel at solving complex tasks by breaking them down into intermediate steps. Several reasoning techniques have driven such improvements. For example, Chain-of-Thought (CoT) prompting \cite{wei2022chain}, generating structured sequences of reasoning steps, has been shown to enhance logical inference. OpenAI’s o1 \cite{jaech2024openai} introduces inference-time scaling for long-COT reasoning, while DeepSeekR1 \cite{guo2025deepseek} enhances reasoning performance using reinforcement learning. 
Along with technical advances, open-source frameworks such as LLaVA \cite{liu2024visual} and Qwen-VL \cite{bai2023qwen} have spurred progress in these areas. These frameworks have been further extended to the medical domain, resulting in various LMMs for medical images and texts. For instance, LLaVA-Med \cite{li2024llava_med} adapts LLaVA’s architecture for biomedical imaging by leveraging figure-caption datasets from PubMed Central. MedTrinity-25M \cite{xie2024medtrinity} builds a comprehensive knowledge base and employs retrieval-augmented generation, using identified regions of interest, such as bounding boxes and segmentation masks, to produce multi-granular textual descriptions. Quilt-LLaVA \cite{seyfioglu2024quilt} constructs image-caption pairs from YouTube videos, while CPath-Omni \cite{sun2024cpath} extends LLaVA to both patch-level and whole-slide image (WSI) level analysis. 

%Newer LLMs have demonstrated superior reasoning abilities compared to their predecessors.

%By leveraging tutorial videos that guide pathologists through the diagnostic process, our approach enables an AI agent to mimic a step‐by‐step diagnostic reasoning process, ultimately arriving at a final diagnosis in a more natural and interpretable manner. 

Most LMMs in the medical domain focus on answering questions based on single images. Though promising, single images can be problematic, especially for diagnostic tasks in pathology. High magnification images miss global structural information and low magnification images lack fine details. 
Alternatively, videos can provide unique and rich sequential visual descriptions, which are incompatible with other sources, such as PubMed articles in PathAsst \cite{sun2024PathAsst}, pathology reports \cite{sun2024cpath}, or even brief Twitter posts in PLIP\cite{huang2023visual}. 
%Educational YouTube videos, in particular, offer rich, sequential visual descriptions that other sources—such as PubMed articles in PathAsst\cite{sun2024PathAsst}, pathology reports \cite{sun2024cpath}, or even brief Twitter posts in PLIP\cite{huang2023visual}—cannot provide. 
Educational YouTube videos are of particular interest since these are publicly available and typically follow a structured pedagogical process: beginning with low-magnification overviews and progressing to high-magnification examinations, they clearly illustrate which features to observe for different organs and diseases\footnote{\scriptsize \url{https://www.youtube.com/watch?v=THhvSJzWEvw}}. This inherent structure makes them an ideal resource for constructing COT reasoning processes in diagnostic tasks, not only boosting model performance but also offering clear insights into the reasoning behind each predicted diagnosis.
Although previous studies, e.g., Quilt-LLaVA \cite{seyfioglu2024quilt}, have employed educational YouTube videos to construct pathology datasets, there is an alignment issue where a single frame represents an entire video segment and its transcription, often exceeding its visual content. For instance, a low-magnification image frame is paired with high-magnification features in text. 

%our video-based data mitigates the alignment issue in Quit-LLaVA \cite{seyfioglu2024quilt}, where a single frame is assumed to match a video segment’s transcription, often exceeding its visual content. For instance, a low-magnification frame may be paired with high-magnification features in text. 

% Moreover, single-frame approaches fail to address whether a region is "too close or too far" for diagnosis.

To address these issues, we propose VideoPath-LLaVA, a diagnostic reasoning model that generates both diagnoses and detailed descriptive explanations for pathology-related videos and images. To the best of our knowledge, this is the first attempt to introduce pathology video understanding. 
To train and evaluate VideoPath-LLaVA, we construct VideoPath-Instruct, a dataset of 4278 curated pathology videos paired with instruction-following Q\&A, which will be made publicly available. 
By leveraging VideoPath-Instruct, VideoPath-LLaVA mimics a step‐by‐step diagnostic reasoning process, ultimately arriving at a final diagnosis in a more natural and interpretable manner. 

%By leveraging VideoPath-LLaVA that guide pathologists through the diagnostic process, VideoPath-LLaVA enables an AI agent to mimic a step‐by‐step diagnostic reasoning process, ultimately arriving at a final diagnosis in a more natural and interpretable manner. 

%The experimental results demonstrate that VideoPsth-LLaVA and VideoPath-Instruct can serve as the basis for building transparent and interpretable decision support systems in pathology.

%Our main contributions can be summarized as follows:
%\begin{itemize}
%    \item VideoPath-Instruct: A publicly available dataset of 4538 curated pathology videos paired with instruction-following Q\&A.
%    \item VideoPath-LLaVA: A diagnostic reasoning model that generates both diagnoses and detailed descriptive explanations from videos and pathology images.
%\end{itemize}

% Herein, we construct a VideoPath-Instruction as an attempt to bring the diagnosis of AI agents with visual reasons. 
% \cite{hu2025video}
% % multi-resolution

%Write something more about the reasoning part,

\section{Method}

\begin{figure*}[!t]
\centering
\includegraphics[width=\textwidth]{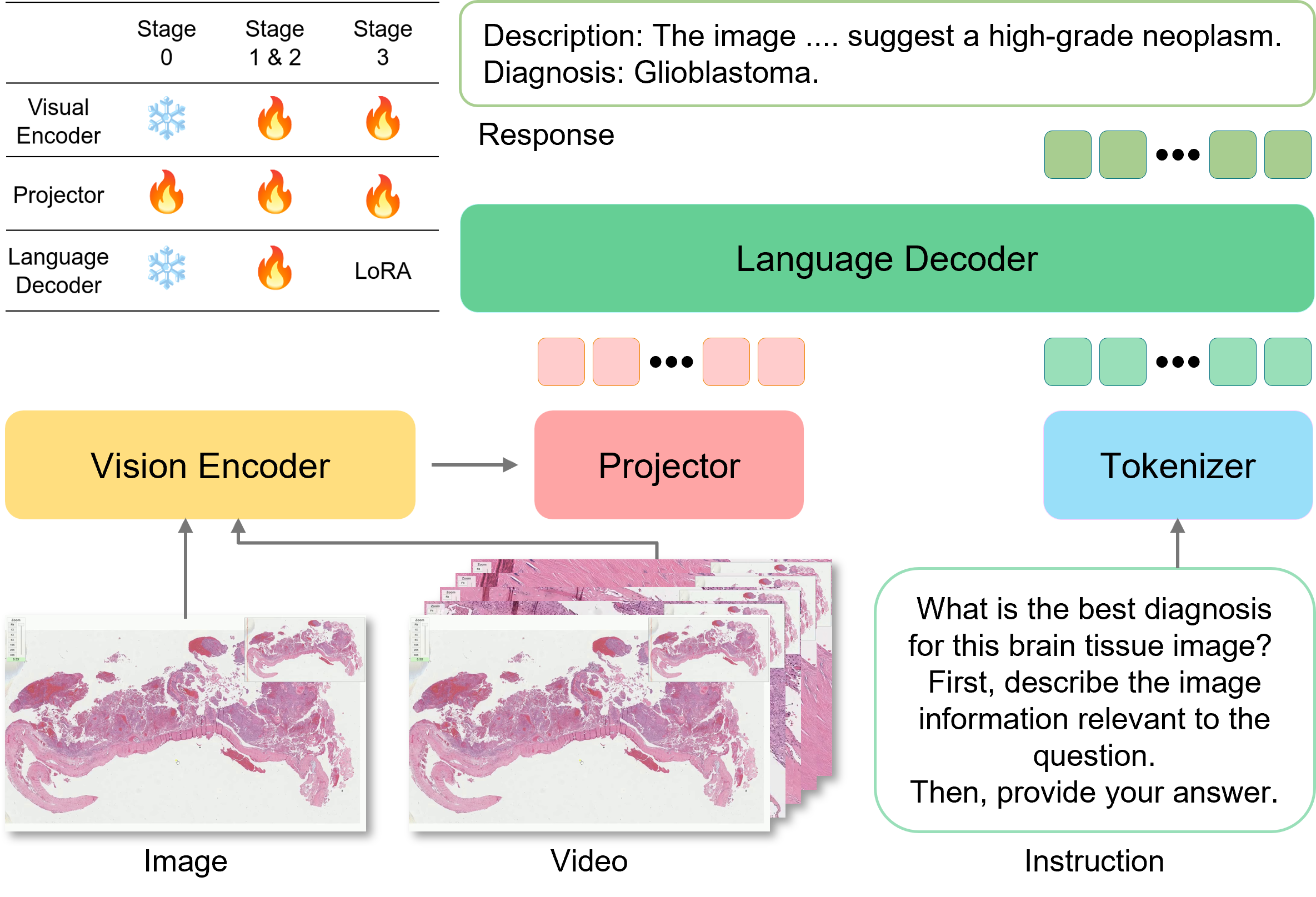}
\caption{Overview of the VideoPath-LLaVA model architecture and its freezing and tuning strategy at each stage. The model comprises three main components: a vision encoder (ViT), a projector, and a large language decoder (LLM).} \label{fig:network}
\end{figure*}

\subsection{Model Architecture}
We build our model on the LLaVA-ov \cite{li2024llava_onevision} architecture with a minor modification: we replace Qwen-2 \cite{yang2407qwen2} with the latest Qwen-2.5 \cite{yang2024qwen2} in the LLM component, as it delivers better benchmark performance. The network architecture, illustrated in Figure \ref{fig:network}, comprises three main components: a vision encoder, a projector, and a language decoder. Given an input pair consisting of an image $x_v$ and language instruction $x_q$. The components operate as follows:
% \begin{itemize}

\noindent\textbf{Vision Encoder (ViT).}  We employ the SigLIP \cite{zhai2023sigmoid} encoder $g_{\psi}(.)$ to extract the image feature $z_v = g(x_v)$.

\noindent\textbf{Projector.} A 2-layer MLP $p_{\theta}(.)$ \cite{liu2024improved} projects the image feature $z_v$ into the word embedding space, resulting in  $h_v = p(z_v)$.

\noindent\textbf{Language Decoder (LLM).} We adopt Qwen-2.5-7B as our LLM $f_{\phi}(.)$ with parameters $\phi$. The LLM receives the projected visual feature $h_v$ along with the tokenized instruction $h_q = \textbf{tokenizer}(x_q)$ as input to generate the output response $x_a = f_{\phi}(h_v, h_q)$. For simplicity, we illustrate a one-turn conversation.
% \end{itemize}

\subsection{Training Strategy}\label{sec:training_strategy}
We train our Video-LLaVA model by adopting the multi-stage strategy from LLaVA-OV\cite{li2024llava_onevision}, which equips LLMs with multimodal capabilities. Each training stage leverages a dedicated dataset (Sec. \ref{section:stage_data}). Additionally, we introduce a fusion stage, \textit{Stage 2: Mixed-SFT} to facilitate the seamless visual task transfer from static images to dynamic video content. Overall, the training procedure is divided into four distinct stages, which are outlined below.

%\begin{itemize}
\noindent\textbf{Stage 0. Alignment.} We pre-train the projector $p_{\theta}(.)$ on image-caption pairs to establish a connection between the two pre-trained components: the LLM $f_{\phi}(.)$ and ViT $g_{\psi}(.)$.

\noindent\textbf{Stage 1. Image-SFT.} In this stage, we fine-tune the entire model, including projector $p_{\theta}(.)$, LLM $f_{\phi}(.)$ and ViT $g_{\psi}(.)$, on image instruction tuning datasets. 

\noindent\textbf{Stage 2: Mixed-SFT.} The model is further fine-tuned on a combination of image and auto-segmented video instruction datasets, to facilitate a smooth transfer of visual task learning from images to videos, enhancing performance on video-related tasks in Stage 3.

\noindent\textbf{Stage 3: Video-SFT.} Finally, we fine-tune the model on our video instruction datasets to enable diagnostic reasoning. Given the smaller size of the manually segmented video dataset compared to the 3 previous stages, we apply LoRA tuning \cite{hu2022lora} to fine-tune the LLM $f_{\phi}(.)$, while projector $p_{\theta}(.)$, and ViT $g_{\psi}(.)$ are fine-tuned without LoRA.

Each stage is trained for one epoch on 8 A6000 GPUs, with batch sizes of 4, 1, 2, and 2, respectively, based on GPU capacity limits.

\subsection{Data Preparation}

We first gather 5,917 raw YouTube videos and apply two temporal segmentation processes, producing two datasets: ClipPath and VideoPath. ClipPath contains automatically segmented pathological clips that may not retain the full diagnostic sequence but provide useful pathology-related content, while VideoPath consists of carefully curated segments in a semi-supervised manner that capture the entire diagnostic reasoning process for an image or WSI. From these segmented videos, we construct instruction-tuning datasets: ClipPath-Instruct and VideoPath-Instruct.
For transcription, we use Whisper-small-en \cite{radford2023robust} for English videos and Whisper-large-translate for 53 non-English videos to generate English subtitles.

% \noindent\textbf{Video Segmentation.}
\noindent\textbf{Visual Data Curation.}
\textit{To create ClipPath}, we design the unsupervised segmentation approach as follows. We use segment captions from Quilt-1M \cite{ikezogwo2024quilt}, which segments videos via FFmpeg\footnote{\scriptsize \url{https://ffmpeg.org/}} keyframe extraction. These keyframes represent points of significant visual change in the raw videos and are used to define candidate segments. By matching these captions to our raw video subtitles, we determine the start and end timestamps of the corresponding segments. This process resulted in 140k pathology-related clips, which were subsequently refined using our \textit{Tissue Detection}. 

\textit{To create VideoPath}, we carefully design a semi-supervised segmentation approach. We first applied AutoShot \cite{zhu2023autoshot} to detect candidate segment boundaries, which were manually refined to ensure that each segment encapsulates a complete diagnostic process. The final timestamps were used to extract the corresponding subtitles. This resulted in 4,036 training videos and 242 testing videos. All segments underwent a cleaning process via our \textit{Tissue Detection}, and for testing videos, an additional \textit{Text Removal} step was applied to prevent text-based leakage.

% To filter low-quality segments, we use GPT-4o-mini \cite{hurst2024gpt} to evaluate transcription relevance and coherence. Each segment receives a quality score (0–5), with only those scoring 3 or higher retained.

\noindent\textbf{Visual Data Refinement.} (1) \textit{Tissue Detection}: Raw videos often include irrelevant elements such as human figures. To refine the data, we manually annotate 5,648 frames (4,538 for training, 1,110 for validation), labeling pathology regions and human figures. We then train a YOLOv10-based pathology detector (YOLO-Path) \cite{wang2025yolov10}, to extract pathology regions while masking human areas by painting them white. (2) \textit{Text Removal}: To prevent large multimodal models (LMMs) from relying on textual clues, we adopt the docTR text recognition model \cite{doctr2021} to detect overlaid diagnostic text. The identified text is removed using inpainting with surrounding pixel information \cite{telea2004image}, ensuring a clean visual dataset.

\noindent\textbf{Instruction Generation.} Inspired by LLaVA \cite{liu2024visual}, we leverage LLMs to construct ClipPath-Instruct and VideoPath-Instruct.

\textit{For ClipPath}, we employ a two-stage prompting approach with LLMs to ensure clarity, relevance, and diagnostic accuracy. Initially, the prompt \textit{"Describe this image in detail."} generates a comprehensive histopathological description. Each raw clip subtitle is evaluated on a 0–5 quality scale based on relevance, sufficiency, and diagnostic adequacy. If it scores $\geq$ 3, the generated description is retained. If the score is $<$3, indicating that the subtitle may either lack sufficient detail or contain noise, we proceed to stage two. At this stage, we apply the alternative prompt, \textit{"Provide a concise description of this image."} to better align with the brevity of the available information. The raw clip subtitle is re-evaluated, and if it meets the quality threshold, the generated concise description is retained; otherwise, it is discarded. This approach yields 140k video Q\&A samples, referred to as \textbf{ClipPath-Instruct}.

\textit{For VideoPath}, we propose a novel approach to generate annotations for VideoPath-Instruct by leveraging LLMs to produce both descriptive and diagnostic outputs from video transcripts. \textbf{In contrast to prior works that typically provide a single question and corresponding instructions for generating visual instruction datasets, our method employs chain-of-thought (CoT) prompting \cite{wei2022chain} to systematically distill the intrinsic reasoning capabilities of LLMs.} Specifically, we engineer our CoT prompts with instructions such as \textit{"What is your diagnosis for this image? First, describe the relevant details, then provide your answer."} This ensures that the LLM first outlines key pathological features before reaching a diagnosis, generating an explicit reasoning chain that serves as supervision for SFT, ultimately improving interpretability and performance. We enhance the reliability of the generated CoT diagnostic reasoning data by integrating zero-shot prompting techniques that guide the LLM in extracting key visual features from the transcript, thereby mitigating the risk of hallucinations.
Here, we obtain 4,036 pathology videos paired with instruction-following Q\&A samples for training and 242 for testing, referred to as \textbf{VideoPath-Instruct.} In our experiments, we employ the GPT-4o-mini LLM \cite{hurst2024gpt} to balance cost and performance.

% To filter low-quality segments, we use GPT-4o-mini \cite{hurst2024gpt} to evaluate transcription relevance and coherence. Each segment receives a quality score (0–5), with only those scoring 3 or higher retained.

%  For testing video, text output docTR\cite{doctr2021} are detected and inpainted.

% This framework not only enriches the interpretability of the model’s predictions but also sets the stage for more robust downstream diagnostic applications.

%  and diagnosis labels
\subsection{Datasets for Multi-Stage Training}\label{section:stage_data}

To perform the training procedure outlined in Sec. \ref{sec:training_strategy} for VideoPath-LLaVA, we also leverage additional datasets beyond ClipPath-Instruct and VideoPath-Instruct, as outlined below.

\noindent\textit{Stage 0: Alignment.} The alignment data comprises image-caption pairs, including 723k samples from Quilt-1M, 223k samples from PathAsst, and 4k bladder pairs \cite{zhang2019pathologist}. Quilt-1M images are preprocessed using our visual data refinement.

\noindent\textit{Stage 1: Image-SFT}, the image language instruction dataset contains 107k samples from Quilt-LLaVA and 100k samples from PathAsst. Quilt-LLaVA images are similarly refined using our visual data refinement.

\noindent\textit{Stage 2: Mixed-SFT}, the model is fine-tuned on a combination of image-based instructions (Stage 1: Image-SFT) and our video-based instructions \textbf{ClipPath-Instruct}. 

% This diverse collection facilitates the seamless transfer of visual instruction capabilities from static images to dynamic video content.

\noindent\textit{Stage 3: Video-SFT}, we further fine-tune the model on our video instruction dataset, \textbf{VideoPath-Instruct}, which comprises 4,036 pairs. This stage is critical for enhancing the model’s performance on video-specific tasks, particularly in diagnostic reasoning. We finally evaluate VideoPath-LLaVA on the testing set of 242 videos.

% Our goal is to effectively leverage the LLM and visual model's pre-trained capabilities and assess the effect of LLaVA pretrained on the big and medical pretrained datasets.

% \subsection{Experiments}
% \subsubsection{Implementation Details}

% \subsubsection{Evaluation} \label{section:Evaluation}

\section{Results}

\begin{table*}[!t]
% \begin{threeparttable}
\centering
% \resizebox{0.9\columnwidth}{!}
{\begin{threeparttable}

% \begin{center}
\caption{Performance benchmarking of text generation models on VideoPath-Instruct. }\label{tab:results}
\setlength{\tabcolsep}{6pt} % Default value: 6pt
\renewcommand{\arraystretch}{1} % Default value: 1
\begin{tabular}{|c|c|c|c|c|c|}
\toprule
	Method & Context & Correct & Detail &  Avg  & Norm-Avg\\
\hline
\rowcolor[HTML]{F5F5FF} 
\multicolumn{6}{|c|}{Open-source LMMs} \\
\hline
LlaVA-OV \cite{li2024llava_onevision} &	1.88 &	1.53 &	1.76 &	1.72 &	34.49 \\
LlaVA-Video  \cite{li2024llava_onevision}  &	2.00 &	1.60 &	1.99 &	1.86 &	37.27 \\
InternVL2-8B \cite{chen2024internvl} &	1.98 &	1.64 &	1.99 &	1.87 &	37.33 \\
Qwen2-VL \cite{Qwen2.5-VL} &	2.14 &	1.79 &	2.06 &	2.00 &	39.94 \\
Qwen2.5-VL \cite{Qwen2.5-VL} &	2.08 &	1.99 &	1.98 &	2.02 &	40.30 \\
\hline
\rowcolor[HTML]{F0F0FF} 
\multicolumn{6}{|c|}{Proprietary LMMs} \\ 
\hline
Gemini-1.5-Pro \cite{google_gemini} &	2.12 &	1.93 &	2.05 &	2.03 &	40.63 \\
Gemini-2.0-Flash \cite{google_gemini} &	2.33 &	2.09 &	2.20 &	2.21 &	44.10 \\
Claude-3.7-Sonnet \cite{anthropic2025claude} &	2.46 &	2.37 &	2.55 &	2.46 &	49.17 \\
GPT-4o \cite{hurst2024gpt}  &	2.69 &	2.69 &	2.36 &	2.58 &	51.60 \\

\hline
 \rowcolor[HTML]{EAEAFF} 
\multicolumn{6}{|c|}{Supervised Fine-Tuning (50\% Video-SFT Data)} \\ 
LLaVA-OV (Baseline) &	1.83 &	1.51 &	2.02 &	1.79 &	35.76 \\
VideoPath-LLaVA (w/o Stage 2)  &	2.56 &	2.49 &	2.55 &	2.53 &	50.63 \\
VideoPath-LLaVA (Ours)&	2.81 &	2.78 &	2.62 &	2.73 &	54.66 \\
\hline
 \rowcolor[HTML]{E0E0FF} 
\multicolumn{6}{|c|}{Supervised Fine-Tuning (Full Video-SFT Data)} \\
LLaVA-OV (Baseline) &	1.86 &	1.40 &	2.03 &	1.76 &	35.21 \\
VideoPath-LLaVA (w/o Stage 2)  &	2.74 &	2.68 &	2.69 &	2.70 &	54.08 \\
VideoPath-LLaVA (Ours)&	\textbf{2.82} &	\textbf{2.82} &	\textbf{2.67 }&	\textbf{2.77} &	\textbf{55.40} \\
    \bottomrule
\end{tabular}
\begin{tablenotes}
\scriptsize
\item[*] All non-proprietary LMMs (7B LLM variants) process 32 frames at \(384 \times 384\) pixels, while Qwen2.5-VL and InternVL2-8B handle only 16 due to memory limits.
\end{tablenotes}
\end{threeparttable}}
\end{table*}

\begin{figure*}[!t]
\centering
\label{fig:prediction}
% \begin{adjustbox}{width=\textwidth}
\scriptsize
% \resizebox{0.9\columnwidth}{!}
{\begin{tabular}{p{0.12\textwidth}p{0.88\textwidth}}
% \noalign{\smallskip}
     \toprule
\multicolumn{2}{c}{\textbf{Visual input example: Visual description and diagnosis}} \\
     \toprule
% \noalign{\smallskip}
\multicolumn{2}{c}{\includegraphics[width=0.9\linewidth]{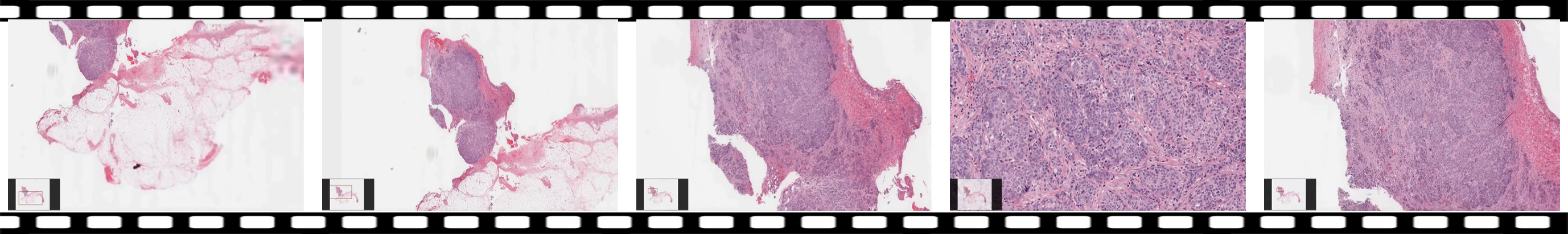}} \\
\hline
\textbf{User} & \textbf{Q}: What is the best diagnosis for the ovary/fallopian tube tissue? First, describe the image's information relevant to the question. Then, provide your answer. \\
& \textbf{A}: The image displays histological sections predominantly composed of malignant epithelial cells, characterized by marked cellular pleomorphism and high mitotic activity. The \textcolor{blue}{nuclei exhibit irregular contours}, with prominent macronucleoli. There is evidence of \textcolor{blue}{atypical mitotic figures} and a \textcolor{blue}{nested architectural pattern} within the tissue. The background may contain \textcolor{blue}{desmoplastic stroma}, which is indicative of invasive behavior. The proliferation of \textcolor{blue}{atypical cells} alongside the \textcolor{blue}{high mitotic count} supports the diagnosis of a high-grade neoplasm, likely of ovarian origin.\\
& Diagnosis: \textbf{High-grade serous carcinoma} \\
\hline
\textbf{GPT-4o} & The images show a histological section of tissue with a dense cellular area. The cells appear to be arranged in \textcolor{blue}{nests and sheets}, with a high nucleus-to-cytoplasm ratio, and there is evidence of \textcolor{blue}{mitotic activity}. The stroma is fibrous and there is a clear demarcation between the tumor and surrounding tissue.
\\
&Based on these features, the best diagnosis for this ovary/fallopian tube tissue is a \textcolor{blue}{"serous carcinoma,"} which is a type of epithelial ovarian cancer characterized by these histological patterns. \\
&Score: (Context/correctness/detail):  4 5 4 (GPT-3.5-turbo-0613) \\
\hline
\textbf{VideoPath-LLaVA (Ours)} & The image shows a histological section of the fallopian tube exhibiting features consistent with serous carcinoma. The epithelium is characterized by a papillary architecture, which is typical for this type of tumor. The cells are arranged in a complex pattern, forming multiple layers and displaying significant \textcolor{blue}{nuclear atypia}. The \textcolor{blue}{nuclei are enlarged and irregularly shaped}, indicating high-grade malignancy. There is also evidence of \textcolor{blue}{increased mitotic activity}, further supporting the diagnosis of a malignant process. The  \textcolor{blue}{stroma appears desmoplastic}, which is often associated with invasive tumors. Overall, the histopathological features observed are indicative of a high-grade serous carcinoma.\\
&Diagnosis: \textcolor{blue}{High-grade serous carcinoma of the fallopian tube.} \\
&Score: (Context/correctness/detail): 5 5 4 (GPT-3.5-turbo-0613) \\
\hline

% Crimson
\end{tabular}}
\caption{Comparison of VideoPath-LLaVA and GPT-4o in the visual reasoning task for diagnosing high-grade serous carcinoma. While both models correctly identify serous carcinoma, GPT-4o fails to recognize key features such as nuclear atypia and desmoplastic stroma, which are critical for assessing tumor invasiveness, leading to a less precise grading of malignancy.}\label{tab:prediction_example}
 % \end{adjustbox}
\end{figure*}

\textbf{Quantitative Results.} Table \ref{tab:results} benchmarks VideoPath-LLaVA against proprietary and open-source LMMs on the VideoPath-Instruct test set. We assess context, correctness, and detail orientation (0–5 scale) using the Video-ChatGPT metric \cite{maaz2023video} and report both average (Avg) and normalized (Norm-Avg, 0–100 scale) scores, with evaluations performed using GPT-3.5-turbo-0613.

\textit{Comparison with Open-Source and Proprietary Models.}
Among open-source LMMs, Qwen2.5-VL achieves the highest performance (Avg: 2.02, Norm-Avg: 40.30), followed by Qwen2-VL (Avg: 2.00, Norm-Avg: 39.94). LLaVA-Video and InternVL2-8B also demonstrate competitive results, (Avg: 1.86, 1.87).

For proprietary LMMs, Claude-3.7-Sonnet and GPT-4o significantly outperform open-source alternatives, with GPT-4o achieving the highest Avg (2.58) and Norm-Avg (51.60). Gemini-2.0-Flash also demonstrates strong performance (Avg: 2.21, Norm-Avg: 44.10), surpassing Gemini-1.5-Pro.

\textit{SFT VideoPath-LLaVA on VideoPath-Instruct.}
We compare our VideoPath-LLaVA with two baselines: (1) LLaVA-OV (Baseline): SStandard SFT of LLaVA-OV on VideoPath-Instruct; (2) VideoPath-LLaVA (w/o Stage 2): A pathology-specific model trained in three sequential stages, Alignment, Image-SFT, and Video-SFT, without Stage 2: Mixed-SFT; and (3) VideoPath-LLaVA (Ours): Extends training with Stage 2: Mixed-SFT and applies LoRA tuning to the LLM in Stage 3 to enhance video-based diagnostic reasoning.

VideoPath-LLaVA (w/o Stage 2) markedly improves over the baseline, increasing the average score from 1.76 to 2.70 and the normalized score from 35.21 to 54.08. Adding Stage 2 in VideoPath-LLaVA (Ours) further enhances diagnostic reasoning, achieving the highest scores (Avg: 2.77, Norm-Avg: 55.40), even surpassing GPT-4o despite using a smaller Qwen2.7-7B LLM.

To assess robustness under data limitations, we fine-tuned models with only 50\% of the video training data. In this setting, VideoPath-LLaVA (Ours) maintains strong performance (Avg: 2.73, Norm-Avg: 54.66), outperforming both baselines. These results highlight the effectiveness of pathology-specific multimodal instruction tuning, particularly Stage 2: Mixed-SFT, in improving diagnostic reasoning for pathology video analysis.

\noindent\textbf{Qualitative Results.}
To further illustrate our findings, we present a qualitative comparison of VideoPath-LLaVA and GPT-4o on a representative example from the test set. This example reflects the general trends observed in our quantitative evaluation (Table \ref{tab:results}), where VideoPath-LLaVA outperforms GPT-4o in identifying key histopathological features and diagnostic accuracy.

\noindent\textbf{Comparison between LoRA vs. Full Fine-tuning.}
Table \ref{tab:lora} compares LoRA tuning LLM with full fine-tuning LLM in the final SFT stage. Prior studies report mixed findings: VILA \cite{lin2024vila} shows fine-tuning superiority, while LlamaFactory \cite{zheng2024llamafactory} finds LoRA slightly better. Given our smaller video SFT dataset, LoRA tuning LLM proves advantageous, improving average scores for both VideoPath-LLaVA and VideoPath-LLaVA (w/o Stage 2), increasing from 2.75 to 2.77 and from 2.70 to 2.74, respectively.

% \begin{figure*}[!t]
% \centering

\begin{table*}[!t]
\centering

\caption{LoRA tuning outperforms fine-tuning LLM on 2 out of 3 settings.}\label{tab:lora}
\setlength{\tabcolsep}{3pt} % Default value: 6pt
\renewcommand{\arraystretch}{1} % Default value: 1
\resizebox{0.9\textwidth}{!}{ % Resize to 90% of text width
\begin{tabular}{|c|c|c|c|c|c|c|c|}
     \toprule
% \multicolumn{2}{|c|}{	Method} & Context & Correct & Detail &  Avg  & Norm-Avg\\
Tuning & Method & Context & Correct & Detail &  Avg  & Norm-Avg\\
     \hline

 &	LLaVA-ov (Baseline) &	1.86 &	1.40 &	2.03 &	1.76 &	\textbf{35.21} \\
Full Fine-tune &	VideoPath-LLaVA (w/o Stage 2) &	2.74 &	2.68 &	2.69 &	2.70 &	54.08 \\
 &	VideoPath-LLaVA &	2.78 &	2.79 &	2.67 &	2.75 &	54.93 \\
 \hline
&	LLaVA-OV (Baseline) &	1.79 &	1.34 &	2.01 &	1.72 &	34.33 \\
LoRA   &	 VideoPath-LLaVA (w/o Stage 2) &	2.73 &	2.77 &	2.72 &	2.74 &	\textbf{54.79} \\
(r=128) &	VideoPath-LLaVA (Ours)&	2.82 &	2.82 &	2.67 &	2.77 &	\textbf{55.40} \\

    \bottomrule
\end{tabular}
}
\end{table*}

\section{Conclusion}
We introduce VideoPath-LLaVA and VideoPath-Instruct, representing the first large multimodal model and pathology video instruction-following dataset in the field. Our model delivers accurate diagnoses while identifying key histopathological features through chain-of-thought (CoT) reasoning, thereby enhancing transparency and interpretability. VideoPath-LLaVA has the potential to enhance clinical decision support systems by delivering prompt, comprehensive diagnostic insights and leveraging multi-frame analysis to support accurate decision-making. 

\noindent\textbf{Limitations \& Future Work.} Nevertheless, the lack of human validation and reliance on YouTube-sourced data pose challenges that warrant further investigation. Future work will focus on dataset expansion, performance enhancement, and expert validation to improve clinical applicability and generalizability, particularly for rare pathologies.

\bibliographystyle{splncs04} % makes bibtex use spiebib.bst

\begin{thebibliography}{10}
\providecommand{\url}[1]{\texttt{#1}}
\providecommand{\urlprefix}{URL }
\providecommand{\doi}[1]{https://doi.org/#1}

\bibitem{anthropic2025claude}
Anthropic: Claude 3.7 sonnet system card (2025), \url{https://assets.anthropic.com/m/785e231869ea8b3b/original/claude-3-7-sonnet-system-card.pdf}, accessed: February 26, 2025

\bibitem{bai2023qwen}
Bai, J., Bai, S., Yang, S., Wang, S., Tan, S., Wang, P., Lin, J., Zhou, C., Zhou, J.: Qwen-vl: A frontier large vision-language model with versatile abilities. arXiv preprint arXiv:2308.12966  (2023)

\bibitem{Qwen2.5-VL}
Bai, S., Chen, K., Liu, X., Wang, J., Ge, W., Song, S., Dang, K., Wang, P., Wang, S., Tang, J., Zhong, H., Zhu, Y., Yang, M., Li, Z., Wan, J., Wang, P., Ding, W., Fu, Z., Xu, Y., Ye, J., Zhang, X., Xie, T., Cheng, Z., Zhang, H., Yang, Z., Xu, H., Lin, J.: Qwen2.5-vl technical report. arXiv preprint arXiv:2502.13923  (2025)

\bibitem{chen2024internvl}
Chen, Z., Wu, J., Wang, W., Su, W., Chen, G., Xing, S., Zhong, M., Zhang, Q., Zhu, X., Lu, L., et~al.: Internvl: Scaling up vision foundation models and aligning for generic visual-linguistic tasks. In: Proceedings of the IEEE/CVF Conference on Computer Vision and Pattern Recognition. pp. 24185--24198 (2024)

\bibitem{google_gemini}
Google: Gemini (nd), \url{https://ai.google.dev/gemini-api/docs/models/gemini}, accessed: February 26, 2025

\bibitem{guo2025deepseek}
Guo, D., Yang, D., Zhang, H., Song, J., Zhang, R., Xu, R., Zhu, Q., Ma, S., Wang, P., Bi, X., et~al.: Deepseek-r1: Incentivizing reasoning capability in llms via reinforcement learning. arXiv preprint arXiv:2501.12948  (2025)

\bibitem{hu2022lora}
Hu, E.J., Shen, Y., Wallis, P., Allen-Zhu, Z., Li, Y., Wang, S., Wang, L., Chen, W., et~al.: Lora: Low-rank adaptation of large language models. ICLR  \textbf{1}(2), ~3 (2022)

\bibitem{huang2023visual}
Huang, Z., Bianchi, F., Yuksekgonul, M., Montine, T.J., Zou, J.: A visual--language foundation model for pathology image analysis using medical twitter. Nature medicine  \textbf{29}(9),  2307--2316 (2023)

\bibitem{hurst2024gpt}
Hurst, A., Lerer, A., Goucher, A.P., Perelman, A., Ramesh, A., Clark, A., Ostrow, A., Welihinda, A., Hayes, A., Radford, A., et~al.: Gpt-4o system card. arXiv preprint arXiv:2410.21276  (2024)

\bibitem{ikezogwo2024quilt}
Ikezogwo, W., Seyfioglu, S., Ghezloo, F., Geva, D., Sheikh~Mohammed, F., Anand, P.K., Krishna, R., Shapiro, L.: Quilt-1m: One million image-text pairs for histopathology. Advances in neural information processing systems  \textbf{36} (2024)

\bibitem{jaech2024openai}
Jaech, A., Kalai, A., Lerer, A., Richardson, A., El-Kishky, A., Low, A., Helyar, A., Madry, A., Beutel, A., Carney, A., et~al.: Openai o1 system card. arXiv preprint arXiv:2412.16720  (2024)

\bibitem{li2024llava_onevision}
Li, B., Zhang, Y., Guo, D., Zhang, R., Li, F., Zhang, H., Zhang, K., Zhang, P., Li, Y., Liu, Z., et~al.: Llava-onevision: Easy visual task transfer. arXiv preprint arXiv:2408.03326  (2024)

\bibitem{li2024llava_med}
Li, C., Wong, C., Zhang, S., Usuyama, N., Liu, H., Yang, J., Naumann, T., Poon, H., Gao, J.: Llava-med: Training a large language-and-vision assistant for biomedicine in one day. Advances in Neural Information Processing Systems  \textbf{36} (2024)

\bibitem{lin2024vila}
Lin, J., Yin, H., Ping, W., Molchanov, P., Shoeybi, M., Han, S.: Vila: On pre-training for visual language models. In: Proceedings of the IEEE/CVF Conference on Computer Vision and Pattern Recognition. pp. 26689--26699 (2024)

\bibitem{liu2024improved}
Liu, H., Li, C., Li, Y., Lee, Y.J.: Improved baselines with visual instruction tuning. In: Proceedings of the IEEE/CVF Conference on Computer Vision and Pattern Recognition. pp. 26296--26306 (2024)

\bibitem{liu2024visual}
Liu, H., Li, C., Wu, Q., Lee, Y.J.: Visual instruction tuning. Advances in neural information processing systems  \textbf{36} (2024)

\bibitem{maaz2023video}
Maaz, M., Rasheed, H., Khan, S., Khan, F.S.: Video-chatgpt: Towards detailed video understanding via large vision and language models. arXiv preprint arXiv:2306.05424  (2023)

\bibitem{doctr2021}
Mindee: doctr: Document text recognition. \url{https://github.com/mindee/doctr} (2021)

\bibitem{radford2023robust}
Radford, A., Kim, J.W., Xu, T., Brockman, G., McLeavey, C., Sutskever, I.: Robust speech recognition via large-scale weak supervision. In: International conference on machine learning. pp. 28492--28518. PMLR (2023)

\bibitem{seyfioglu2024quilt}
Seyfioglu, M.S., Ikezogwo, W.O., Ghezloo, F., Krishna, R., Shapiro, L.: Quilt-llava: Visual instruction tuning by extracting localized narratives from open-source histopathology videos. In: Proceedings of the IEEE/CVF Conference on Computer Vision and Pattern Recognition. pp. 13183--13192 (2024)

\bibitem{sun2024cpath}
Sun, Y., Si, Y., Zhu, C., Gong, X., Zhang, K., Chen, P., Zhang, Y., Shui, Z., Lin, T., Yang, L.: Cpath-omni: A unified multimodal foundation model for patch and whole slide image analysis in computational pathology. arXiv preprint arXiv:2412.12077  (2024)

\bibitem{sun2024PathAsst}
Sun, Y., Zhu, C., Zheng, S., Zhang, K., Sun, L., Shui, Z., Zhang, Y., Li, H., Yang, L.: Pathasst: A generative foundation ai assistant towards artificial general intelligence of pathology. In: Proceedings of the AAAI Conference on Artificial Intelligence. vol.~38, pp. 5034--5042 (2024)

\bibitem{telea2004image}
Telea, A.: An image inpainting technique based on the fast marching method. Journal of graphics tools  \textbf{9}(1),  23--34 (2004)

\bibitem{wang2025yolov10}
Wang, A., Chen, H., Liu, L., Chen, K., Lin, Z., Han, J., et~al.: Yolov10: Real-time end-to-end object detection. Advances in Neural Information Processing Systems  \textbf{37},  107984--108011 (2025)

\bibitem{wei2022chain}
Wei, J., Wang, X., Schuurmans, D., Bosma, M., Xia, F., Chi, E., Le, Q.V., Zhou, D., et~al.: Chain-of-thought prompting elicits reasoning in large language models. Advances in neural information processing systems  \textbf{35},  24824--24837 (2022)

\bibitem{xie2024medtrinity}
Xie, Y., Zhou, C., Gao, L., Wu, J., Li, X., Zhou, H.Y., Liu, S., Xing, L., Zou, J., Xie, C., et~al.: Medtrinity-25m: A large-scale multimodal dataset with multigranular annotations for medicine. arXiv preprint arXiv:2408.02900  (2024)

\bibitem{yang2407qwen2}
Yang, A., Yang, B., Hui, B., Zheng, B., Yu, B., Zhou, C., Li, C., Li, C., Liu, D., Huang, F., et~al.: Qwen2 technical report, 2024. URL https://arxiv. org/abs/2407.10671  (2024)

\bibitem{yang2024qwen2}
Yang, A., Yang, B., Zhang, B., Hui, B., Zheng, B., Yu, B., Li, C., Liu, D., Huang, F., Wei, H., et~al.: Qwen2. 5 technical report. arXiv preprint arXiv:2412.15115  (2024)

\bibitem{zhai2023sigmoid}
Zhai, X., Mustafa, B., Kolesnikov, A., Beyer, L.: Sigmoid loss for language image pre-training. In: Proceedings of the IEEE/CVF International Conference on Computer Vision. pp. 11975--11986 (2023)

\bibitem{zhang2019pathologist}
Zhang, Z., Chen, P., McGough, M., Xing, F., Wang, C., Bui, M., Xie, Y., Sapkota, M., Cui, L., Dhillon, J., et~al.: Pathologist-level interpretable whole-slide cancer diagnosis with deep learning. Nature Machine Intelligence  \textbf{1}(5),  236--245 (2019)

\bibitem{zheng2024llamafactory}
Zheng, Y., Zhang, R., Zhang, J., Ye, Y., Luo, Z., Feng, Z., Ma, Y.: Llamafactory: Unified efficient fine-tuning of 100+ language models. arXiv preprint arXiv:2403.13372  (2024)

\bibitem{zhu2023autoshot}
Zhu, W., Huang, Y., Xie, X., Liu, W., Deng, J., Zhang, D., Wang, Z., Liu, J.: Autoshot: A short video dataset and state-of-the-art shot boundary detection. In: Proceedings of the IEEE/CVF Conference on Computer Vision and Pattern Recognition. pp. 2238--2247 (2023)

\end{thebibliography}

\end{document}